# A Comprehensive Approach to Misspelling Correction with BERT and Levenshtein Distance

Amirreza Naziri[a,b], Hossein Zeinali[a,b,*]

[a]*Computer Engineering Department, Amirkabir University of Technology, Tehran, Iran*
[b]*Sharif DeepMine Ltd., Tehran, Iran*

**Abstract**

Writing, as an omnipresent form of human communication, permeates nearly every aspect of contemporary life. Consequently, inaccuracies or errors in written communication can lead to profound consequences, ranging from financial losses to potentially life-threatening situations. Spelling mistakes, among the most prevalent writing errors, are frequently encountered due to various factors. This research aims to identify and rectify diverse spelling errors in text using neural networks, specifically leveraging the Bidirectional Encoder Representations from Transformers (BERT) masked language model. To achieve this goal, we compiled a comprehensive dataset encompassing both non-real-word and real-word errors after categorizing different types of spelling mistakes. Subsequently, multiple pre-trained BERT models were employed.

To ensure optimal performance in correcting misspelling errors, we propose a combined approach utilizing the BERT masked language model and Levenshtein distance. The results from our evaluation data demonstrate that the system presented herein exhibits remarkable capabilities in identifying and rectifying spelling mistakes, often surpassing existing systems tailored for the Persian language. Notably, the proposed method achieved an increase in the relative F1-score by more than 28 %.

## 1. Introduction

Throughout history, humans have tirelessly endeavored to discover effective means of communication and message conveyance. Over time, they have successfully developed diverse methods of expression, including spoken language, visual artistry through drawing, and the written word. However, it did not take long for writing to emerge as one of the most paramount forms of communication. This is primarily due to the advent of books and other written documents, which allowed for an extensive range of ideas and concepts to be recorded.

The significance of writing cannot be overstated, as it serves as the foundation for the existence and preservation of crucial elements within society. News reports, legal statutes, and educational resources in schools and universities - all owe their existence to the mastery of writing skills. Countless other facets of life are also indebted to this profound ability.

In today's digital age, particularly with the rise of the Internet and social media platforms, virtually anyone can effortlessly produce content. Consequently, it is only natural that the likelihood of encountering writing errors in texts has significantly increased. To illustrate this point further, let us consider the Persian language: two words - "حیات" (meaning "life") and "حیاط" (meaning "yard") - bear striking similarities in terms of both spelling and phonetics. Consequently, individuals may inadvertently utilize one word instead of the intended counterpart.

Additionally, with keyboards and typewriters becoming commonplace tools for written communication, a new set of challenges has arisen. The adjacency between different letter combinations on these devices has amplified the potential for spelling errors compared to traditional handwritten forms. Thus, we find ourselves faced with an ever-present possibility of encountering such typos.

Furthermore, given that writing permeates various critical contexts in our lives, its influence becomes even more pronounced. The transmission of messages can be distorted or manipulated due to errors or intentional misrepresentation within written works. This can lead to false promotion or misinterpretation within scientific realms and other domains.

On the other hand, the presence of incorrect words within a sentence can significantly compromise the quality of results produced by different language models, such as Name Entity Recognition (NER) [1], Semantic Role Labeling (SRL) [2], Next Word Prediction [3], Language Translation models [4], and others. Therefore, it is primarily due to the fact that all these aforementioned models heavily depend on word probabilities in order to effectively carry out their intended tasks. Consequently, it becomes imperative to develop a viable solution that can effectively mitigate such errors.

---

*Corresponding author.
*Email addresses:* naziriamirreza@gmail.com (Amirreza Naziri), hzeinali@aut.ac.ir (Hossein Zeinali)

In this paper, our primary objective revolves around rectifying different types of spelling mistakes through the utilization of machine learning techniques. Given that spelling mistakes can be categorized into two distinct groups, real and non-real-words. Non-real-words refer to those terms that are not typically included within a language's vocabulary. For instance, consider the word "ثبات" (meaning "steadiness") which may be erroneously written as "صبات" due to various reasons, including author negligence. Since this particular word does not exist within the Persian language's lexicon, it becomes relatively straightforward to identify its incorrectness within a given text. However, suggesting alternative words poses a significant challenge due to factors such as an extensive set of candidate words.

Conversely, the second category consists of real-words. However, unlike their non-real counterparts, these words do exist within a language's vocabulary; however, they should not be used in a sentence based on their semantic meaning. For instance, in the phrase "صوت و تصویر" (meaning "sound and vision") substituting "سوت" (meaning "whistle") for "صوت" (meaning "sound") would be considered incorrect. Recognizing this category of words becomes more arduous than identifying non-real-words; nevertheless, owing to the limited number of potential replacement words available for consideration, suggesting an appropriate substitute becomes relatively easier.

To achieve this objective, both the minimum edit distance algorithm [5] and various versions of BERT masked language models [6] were applied. These methodologies are employed in order to effectively address the identified problem and ultimately enhance the accuracy of spelling correction within textual data. The BERT language model is a prominent natural language processing (NLP) model that is used for addressing a wide range of NLP challenges, including but not limited to emotion recognition, question answering, summarization, and translation [7].

In our study, we tried to improve the effectiveness and efficiency of misspelling correction using BERT. Pre-trained BERT masked language models was used to suggest potential words for misspelled words and the minimum edit distance was utilized to identify related words. In addition, the algorithm was refined by incorporating additional heuristic techniques to handle non-real-word and real-word errors more accurately. Moreover, a novel biasing mechanism was introduced to increase the precision of our error correction model.

This paper adopts a structured approach to present its findings. It begins with a discussion of related works in Section 2, where existing literature and research relevant to the topic are examined. Section 3 focuses on data preparation, outlining the procedures employed to collect, clean, and preprocess the data utilized in the study. The proposed method is introduced in Section 4, presenting novel approaches developed by us to address misspelling correction. Subsequently, Section 5 presents the experimental setup, encompassing variables, parameters, and the results obtained from applying the proposed method to the prepared data. The study then delves into error interpretation, analysis, and limitations in Section 6. Finally, the Conclusions section provides a summary of all the findings.

## 2. Related Works

### 2.1. Traditional Methods

Traditional methods for spelling error detection and correction often rely on n-gram [8, 9, 10] and Bayesian approaches [11, 12, 13]. One such system, presented in [14], focuses on English and offers an automated approach to identify and rectify spelling mistakes. This system utilizes the Bayesian model with trigram model. The experimental results demonstrate that this approach attains an accuracy ranging from 86.16% to 89.83%.

Similarly, Mays et al. proposed a comparable model in [15] that utilizes tri-grams to detect approximately 76% of simple spelling errors while successfully correcting 73% of them.

Furthermore, CloniZER [16] aims to develop a language-independent error correction system capable of adapting to any given language for the purpose of rectifying non-real-words. This system employs a triple search tree data structure and leverages non-deterministic traversal techniques to determine the appropriate form for incorrectly spelled words. After conducting 21 scans, CloniZER obtains an accuracy rate exceeding 80%.

Church and Gale [17] proposed a context-based system that incorporates a noisy channel model utilizing a basic word bi-gram model along with frequency estimation techniques.

In another work [18], Carlson et al. demonstrated the effectiveness of n-gram models in correcting spelling errors, achieving an impressive accuracy rate of 92.4% for detecting insertion errors using 5-grams.

Furthermore, a project focusing exclusively on the English language [19] addresses the recognition and correction of real-words exclusively. This method applies an enhanced version of the longest common sequence algorithm, resulting in a recall value of 89% for error detection and 76% for error correction.

A misspelling detection and correction system called Farsi-Spell [20] has been developed specifically for the Persian language, utilizing a large monolingual dataset. The system employs the Mean reciprocal rank (MRR) metric method to evaluate, sort, and assign value to suggested words. Remarkably, this method has achieved an impressive accuracy rate of 94.3% in detecting erroneous words and 67.4% in suggesting the correct alternatives. However, it is important to note that Farsi-Spell is limited to identifying and correcting non-real-word errors.

In another related study by Faili and Azadnia [21], they proposed a context-sensitive spelling checker that attains an F1 score of 80%. This approach incorporates the Bayesian method to enhance its performance. Additionally, the mutual information matrix method is employed to establish meaningful relationships between words.

*2.2. Neural Network Based Methods*

The detection and correction of spelling errors can be effectively achieved by using sequential neural networks [22, 23]. Lee and Kim [24] also achieved remarkable correction performance on English by utilizing an auto encoding (AE) language model-based approach.

In the context of the Arabic language, Alkhatib et al. [25] introduced a novel approach that outperformed Microsoft Word 2013 and Open Office Ayaspell 3.4. Their method relied on a bidirectional long short-term memory mechanism combined with word embedding techniques. Additionally, AraSpell [26] is another noteworthy Arabic spelling correction system that employs various seq2seq model architectures such as Recurrent Neural Network (RNN) [27, 28] and Transformers [29].

Furthermore, similar research has been conducted on the Malaysian language [30], where a long short-term memory (LSTM) network [31] was employed. Moreover, HINDIA [32], a deep learning-based model for spell-checking in Hindi language, exhibited superior performance compared to existing Malaysian spell checkers.

In another study, Etoori et al. [33] proposed SCMIL, an architecture based on neural networks, which demonstrated an impressive accuracy of 85.4% for Hindu and 89.3% for Telugu languages.

Kuznetsov and Urdiales [34] proposed a spelling correction method for different languages using a denoising transformer specifically designed for short strings like queries. This approach proved to be effective in achieving accurate corrections.

For the Azerbaijani language, Ahmadzade and Malekzadeh [35] introduced a sequence-to-sequence model incorporating an attention mechanism. Their model achieved impressive F1 scores of 75% for edit distance 0, 90% for edit distance 1, and 96% for edit distance 2.

Another notable example is the work conducted by Jing et al. [36], who introduced an innovative approach that employs BERT for spelling correction on the English language. This particular research served as the foundation for our own project. The BERT model is used to accurately predict masked words. In order to achieve this, a dataset that encompasses both Cambridge and First English Certificate datasets is utilized. The working methodology involves sequentially masking words within each sentence and subsequently utilizing BERT's language model to estimate the masked words.

During this stage, it explored two distinct approaches. The first approach entails employing the Levenshtein distance [37] to evaluate the words suggested by BERT after its usage. Conversely, the second approach involves utilizing the Levenshtein distance to generate candidate words prior to leveraging BERT. The experimental results demonstrate that the introduced model achieves an impressive accuracy rate of 78.84% when implementing the first approach. However, in the second scenario, if all incorrectly spelled words possess a Levenshtein distance of at least 2 and are present within the dataset, the model attains an even higher accuracy rate of 84.91%.

Overall, these studies highlight the effectiveness of utilizing neural networks in detecting and correcting spelling errors across various languages. The advancements made in each specific language demonstrate the potential of these models in improving spell-checking accuracy and efficiency.

## 3. Dataset Preparation

The dataset used in this research paper is a subset extracted from the digital books available in Taaghche[1] database as well as Persian Wikipedia[2] articles. The dataset consists of entirely accurate and unprocessed data, without any labels or metadata. Consequently, in order to employ this dataset for both training and evaluating the model proposed in this paper, it is necessary to perform pruning and error generation steps. Prior to that, it is advisable to study the characteristics of misspelled words and the methods employed for generating them in Persian language.

*3.1. Different Categories of Misspelling Errors in General*

In general, as discussed in the introduction, there are two main types of spelling errors found in all languages. The first type consists of real-words that exist within the language's general vocabulary; however, these words do not convey the intended meaning, rendering the sentence incomplete. On the other hand, the second category consists of non-real-words that are not listed in any dictionary. Consequently, these words are fundamentally incorrect.

For the purpose of this paper, it was imperative to detect both types of errors. Throughout the project, a comprehensive Persian dictionary containing more than 264K words was used. This dictionary had been meticulously arranged alphabetically and prepared well in advance. Furthermore, each error type was further subdivided into three distinct categories which will be thoroughly examined in the subsequent section.

*3.2. Different Types of Misspelling Errors in Each Category*

**Keyboard Error:** This particular type of error typically occurs as a result of carelessness and inaccuracy when using the keyboard. For instance, the user might have made an error while typing the word "تها" (meaning "alone") and mistakenly wrote it as "دنها" or "منها" (meaning "minus"). In this case, the user has unintentionally pressed the "د" or "م" button instead of pressing the correct "ت" key.

To generate such erroneous words, it is first necessary to establish a one-to-many data structure for storing neighboring letters based on the Persian standard keyboard layout. For exam-ple, the letter "ی" corresponds to a list of letters including "ب، ق، ث، ص، س، ط، ز، ر". Then, one letter from the chosen word, which was in the data structure's key list, was randomly selected and then replaced with one of its corresponding letters from the data structure. Finally, if the generated word exists within the language's vocabulary, it is considered a real-word error; otherwise, it is deemed a non-real-word error.

---
[1] https://taaghche.com
[2] https://www.wikipedia.org

**Substitution Error:** Similar to the previous type, this kind of error also arises due to carelessness and inaccuracy when using either a keyboard or traditional handwriting methods. For instance, while writing the word "بسته" (meaning "package"), the user mistakenly writes it as "سبته". In this case, the user inadvertently interchanged the letters "س" and "ب".

To generate words with substitution errors like these, it is sufficient to randomly select two adjacent letters within a word and swap them. Similarly, in this case, the generated words will be added to the list of real and non-real-word errors.

**Homophone Error:** These errors typically arise from a lack of writing proficiency or the user's imprecision. For instance, while attempting to write the word "ساعد" (meaning "forearm"), the user might mistakenly write it as "صاعد" (meaning "ascending"). In this case, the error occurs due to the similar pronunciation of the Persian letters "س" and "ص".

To generate such words, it is essential to first establish a one-to-many data structure that includes consonants for each letter. For example, the letter "س" corresponds to a list of letters including "ث" and "ص". Next, within a chosen word, one of the letters of the chosen word, which was in this data structure's key list is randomly selected and replaced with one of its consonants. Lastly, if the resulting word exists in the language's vocabulary, it is considered a real-word; otherwise, it is regarded as a non-real-word. These erroneous words are also referred to as polymorphism errors when they are not-real-word errors.

*3.3. Pruning Dataset*

The pruning step was carried out line by line. First, all numeric values, punctuation marks, and Latin letters were removed from the data. Then, the sentences were converted into a list of tokens. If any token was not found in the main vocabulary or if the number of tokens was less than five or exceeded 256, the sentence was excluded.

*3.4. Error Generation*

In the subsequent phase, 50% of the lines remained unaltered while the remaining 50% underwent modifications according to the following guidelines:

If it was feasible to introduce a homophone error within a sentence, there was an 80% likelihood that the word would be replaced with a real-word homophone error.

Table 1: The statistics of the generated training dataset.

| | |
|---|---|
| Total number of sentences | 8,059,076 |
| Total number of correct sentences | 4,775,776 |
| Total number of sentences with Real-word Homophone error | 102,420 |
| Total number of sentences with Real-word Keyboard error | 792,443 |
| Total number of sentences with Real-word Substitution error | 731,165 |
| Total number of sentences with Non-real-word Homophone error | 529,962 |
| Total number of sentences with Non-real-word Keyboard error | 530,321 |
| Total number of sentences with Non-real-word Substitution error | 530,479 |

The remaining data was once again divided into halves (each approximately 25%). The first part was further divided into two equal probability groups (0.5), where one group represented keyboard errors and the other represented substitution errors with real-words. As for the second part, it was divided into three equal probability groups (0.33), comprising keyboard, substitution, and homophone non-real-word errors.

The error generation process was repeated twice on the entire pruned dataset in order to generate training data. Each repetition was conducted separately on the outcome of the pruning step, and subsequently, the results were appended, exact numbers are presented in Table 1.

Since only one randomly selected word in each line became an error during each repetition, there was a low likelihood of repeating incorrect words within the dataset. Consequently, this approach effectively increased the number of erroneous words in the training data.

Moreover, to facilitate the evaluation process, the error generation procedure was conducted once on a subset of 37 thousand lines extracted from the pruned dataset, resulting in the creation of evaluation data. As it can be seen in Table 2, to enable the calculation of metrics based on evaluation outcomes, only those sentences that remained correct and unaltered were retained if they included at least one word that could potentially be a real-word error in another sentence.

**4. Proposed Method**

Generally, the method explained by Jing et al. [36] was used as the baseline in our study. Our aim was to enhance the effectiveness and efficiency of the original approach while addressing certain limitations. The method described as the base-line, provided a solid foundation for our research, as it successfully tackled the problem of misspelling correction using BERT. However, in our research, certain areas were identified where further advancements could be made to achieve even better results.

Table 2: The statistics of the generated evaluation dataset.

| | |
|---|---|
| Total number of sentences | 37040 |
| Total number of sentences with Correct Real-word Homophone | 1022 |
| Total number of sentences with Correct Real-word Keyboard | 5000 |
| Total number of sentences with Correct Real-word Substitution | 5000 |
| Total number of sentences with Real-word Homophone error | 1021 |
| Total number of sentences with Real-word Keyboard error | 5000 |
| Total number of sentences with Real-word Substitution error | 5000 |
| Total number of sentences with Non-real-word Homophone error | 4997 |
| Total number of sentences with Non-real-word Keyboard error | 5000 |
| Total number of sentences with Non-real-word Substitution error | 5000 |

In our research, similar to the baseline method, pre-trained BERT masked language models were applied to suggest potential words for misspelled words. Also, the minimum edit distance was used to identify words closely related to the misspelled ones.

As our first improvement, the algorithm used in the baseline was refined by incorporating additional heuristic techniques. These techniques allowed us to better handle non-real-word and real-word errors, resulting in more accurate and reliable outcomes. Moreover, a novel biasing mechanism that increased the precision of our error correction model was introduced.

Furthermore, the data preparation setup was improved by using a more extensive and diverse dataset with different types of misspelling errors compared to that used in the baseline paper [36]. This expansion enabled us to evaluate our proposed method under various scenarios and validate its robustness for various misspelling errors. Further elaboration on this topic will be provided in the subsequent sections.

### 4.1. Masking Misspelled Words

To use BERT masked language model for suggesting various alternatives for misspelled words, it is crucial to replace the erroneous words with mask tokens, as discussed here [6]. By doing so, BERT can suggest "N" candidate words, which can be adjusted within the range of 1 to the vocabulary size. Another approach is to provide BERT with a list of candidate words, allowing it to rank them based on their relevance within the given sentence.

However, if the provided candidates are not present in BERT's vocabulary, BERT substitutes them with other words, potentially leading to a decrease in result quality [36]. In our study, different versions of biased BERT masked language models were tested and compared to determine the most effective versions for candidate suggestion.

### 4.2. Biasing BERT Masked Language Model

In order to enhance the quality of the best model's results, BERT masked language models were biased using the generated training data. Before delving into the process and its description, it is imperative to address the general procedure of learning a masked language model. Typically, every machine learning model requires a method to assess the deviation between the output and the actual result. This deviation is measured through the cross-entropy cost function [38].

Since the model's output consists of a sequence of words, calculating the cost value for an entire sentence is expensive. Furthermore, research findings [6] indicate that training the model on the complete output does not yield satisfactory outcomes. Consequently, during training, it suffices to apply the cost function solely on a series of specific output words.

Initially, 15% of the overall input data is randomly chosen, with a majority (80%) of these selected words being masked in the input. Thus, the model endeavors to predict these concealed words. However, not all of these words are masked; some are substituted with other random words in the input (10%). In such cases, the model strives to identify less significant words that do not contribute significantly to estimating the output.

The remaining words remain unaltered (10%). Herein lies an attempt by the model to preserve correctly written words within sentences. Nevertheless, as mentioned earlier, these three aforementioned groups collectively constitute only 15% of all input data. Consequently, training can be accomplished with less effort [6].

The initiative presented in this paper, apart from adhering to the aforementioned requirements, incorporates an additional clause into the algorithm. In this scenario, the cost function is also applied to evaluate all incorrectly spelled words within the input. Consequently, the model will prioritize correcting spelling errors.

Furthermore, in case of erroneous input, all three introduced processes can be reapplied; however, it is crucial to emphasize that the output of these words holds significance in all instances. As a result, the model learns to initially disregard these incorrect words and strives to suggest the most suitable candidate words (based on contextual meaning) as accurately as possible.

It is important to note that throughout the entire masking and modification process, it is performed on individual words rather than tokens. This distinction arises due to the fact that a single word may be represented by multiple tokens during tokenization, and it can enhance results [39].

*4.3. Versions of BERT Model*

In order to identify the most suitable model for our problem, various versions of the BERT model were tested. Throughout this project, models implemented by Hooshvare Research Lab[1] were used. These models include different configurations of the BERT model, which have been fine-tuned using Persian datasets. DistilBERT [40], ParsBERT v1.0 [41], ParsBERT v3.0 (i.e. "bert-fa-zwnj-base"), ParsBERT v2.0 (i.e. "bert-fa-based-uncased") were examined to determine the optimal model for our needs.

*4.4. Levenshtein Distance*

The Levenshtein distance, also known as the edit distance, is a metric used to determine the difference between two distinct strings. This metric represents the minimum number of operations (deletion, substitution, and insertion) required to transform one string into another. In essence, the Levenshtein distance shows how similar or dissimilar two strings are [5, 37].

For instance, the edit distance between "خوان" (meaning "tablecloth") and "خامه" (meaning "cream") is 3. This is because three actions are necessary to convert the first word to the second one: removing "و", substituting "ن" instead of "م", and inserting "ه".

There are numerous algorithms for computing the edit distance, with the most renowned being the Needleman-Wench algorithm [42].

*4.5. Approach for Non-real-Word Errors*

To correct non-real-word errors, a specific approach was employed. It involved selecting all words from the dictionary that had an edit distance of one from the misspelled word.

Furthermore, a heuristic approach that found all similar words with an edit distance of two from the misspelled word was implemented. The condition for inclusion is that by only replacing two adjacent letters in candidate words, they can be transformed into the misspelled word.

Then, the masked sentence, along with a list of close words, were fed into BERT's masked language model. The final candidate was selected based on the highest score given by BERT's masked language model.

---

[1] https://huggingface.co/HooshvareLab

*4.6. Approach for Real-word Errors*

To address this type of error, a data structure was created to include all words in the main dictionary. This structure stored all potential candidate words that could be real-word errors caused by shifting, replacing a letter with a neighboring letter on the keyboard, or changing the consonant of the main word.

Subsequently, the masked sentence, along with the list of words in the data structure that contained misspelled words, was fed into BERT's masked language model. If BERT's score for all suggested words fell below a certain adjustable threshold (K), or if the edit distance exceeded 2, the original word was returned. Conversely, if a suggestion met these conditions and had the highest score according to BERT's model, it would be returned as the final candidate. The choice of setting 2 as the edit distance threshold came from BERT's difficulty in handling candidate words that were not present in its dictionary. In such cases, BERT would substitute an unknown candidate with another word, potentially leading to complete inaccuracies. By selecting 2 as the edit distance threshold, erroneous suggestions were eliminated from the final results.

It is important to note that all misspelled words were labeled beforehand; thus, the model only performed correction procedures on desired words.

## 5. Experiments and Results

This section presents the findings for both the initial and biased scenarios of the proposed spelling correction solution. It is important to note that all the results presented in this section were obtained using the evaluation dataset, as detailed in Table 2.

*5.1. Evaluation Criteria*

Throughout the project, a variety of established evaluation criteria were employed to assess the proposed solution. However, due to the relatively low incidence of misspelled words in the text and the predominance of correctly spelled words, the data exhibits an uneven distribution. As a result, traditional metrics such as accuracy may not always be the most appropriate measures in certain cases. This paper utilizes the following metrics for evaluation:

- **True Positive (TP):** This metric represents the number of initially misspelled words that were subsequently corrected by the model.

- **True Negative (TN):** It indicates the number of words that were correctly spelled and remained unchanged by the model.

- **False Negative (FN):** This metric corresponds to the number of words that were initially misspelled and remained uncorrected after passing through the model. It also covers cases where the model recognized the errors but suggested incorrect alternatives.

- **False Positive (FP):** This metric signifies the number of words that were correctly spelled but were mistakenly altered by the model.

- **Accuracy:**

$$Accuracy = \frac{TP + TN}{TP + TN + FP + FN} \quad (1)$$

- **Precision:**

$$Precision = \frac{TP}{TP + FP} \quad (2)$$

- **Recall:**

$$Recall = \frac{TP}{TP + FN} \quad (3)$$

- **F1-Score:**

$$F1\text{--}Score = \frac{2 \times Precision \times Recall}{Precision + Recall} \quad (4)$$

In cases involving non-real-words, precision is the primary metric that can be applied (or accuracy, as both metrics are often considered synonymous in this context). Under such circumstances, if a word was not found in the dataset, assuming that correct sentences were constructed in such a way that all words are part of the core vocabulary, any detection by the model would be classified as an error related to non-real-words. Consequently, the recall index has limited practical relevance in this context, as it typically approximates 100%.

To compute the average of the specified metrics, we employed both Micro and Macro averaging techniques.

- **Micro Average:** This approach computes the desired metric (e.g., accuracy or recall) separately for each class, taking into account the number of samples within each class. The individual metrics are then aggregated and divided by the total sample count. This method proves particularly valuable when handling imbalanced data, as it accommodates variations in sample sizes among different classes.

- **Macro Average:** In contrast, the macro average method calculates the desired metric for each class independently and subsequently computes the average. Unlike the micro average, this approach disregards the sample count within each class, treating all classes with equal weight. In simpler terms, as it does not factor in sample sizes, this method is better suited for balanced data.

Utilizing both micro and macro averaging techniques ensures a comprehensive evaluation of our metrics, while also addressing potential data imbalances.

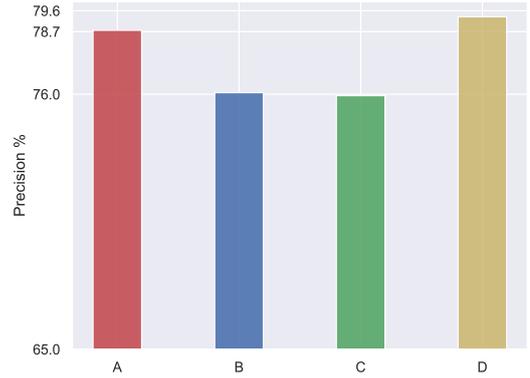

Figure 1: Precision for unbiased models - Non-real-word Errors. (A: ParsBERT v1.0, B: ParsBERT v3.0, C: DistilBERT, D: ParsBERT v2.0)

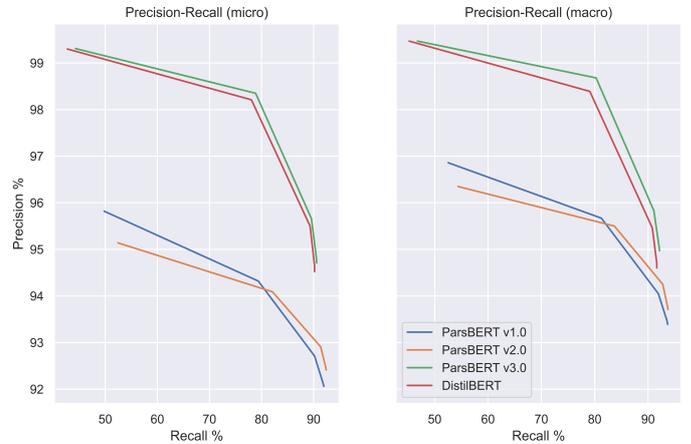

Figure 2: Precision-Recall curve for unbiased models - Real-word Errors.

*5.2. Evaluation Results*

As previously highlighted, the misspelling errors can be categorized into real-word and non-real-word errors, and as such, the results are presented separately for each error category.

Concerning non-real-word errors, the detection and correction of these errors remain consistent across varying thresholds (refer to Section 4.5). Figure 1 demonstrates that both ParsBERT v1.0 and ParsBERT v2.0 exhibit slightly higher precision compared to the other models. Nevertheless, this difference in precision, while noteworthy, does not attain statistical significance, suggesting that all models perform comparably well for non-real-word errors.

For real-word errors, a comprehensive evaluation of all models was conducted using different BERT threshold values (refer to Section 4.6). The thresholds applied in the experiment encompassed values of 1e-1, 1e-3, 1e-5, 1e-7, and 1e-9.

As depicted in Figure 2, the results pertaining to real-word errors can be bifurcated into two distinct groups: one composed of the outcomes for ParsBERT v1.0 and ParsBERT v2.0, and the other consisting of the results for ParsBERT v3.0 and DistilBERT. The former group exhibited higher recall, while the latter demonstrated superior precision across the same thresholds. It is evident that models within each group yielded similar re-sults. Consequently, for the sake of simplicity, one model from each group was chosen for further evaluation. Consequently, ParsBERT v2.0 and ParsBERT v3.0 were selected as representatives from their respective groups, as they consistently outperformed the other models within their categories.

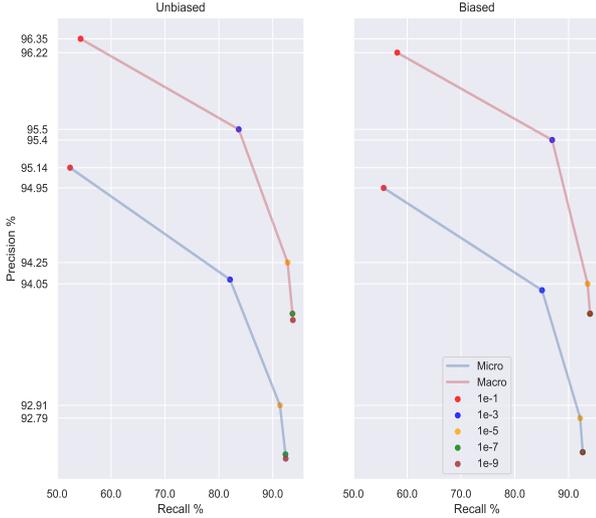

Figure 3: Precision-Recall curve for ParsBERT v2.0 - Real-word Errors.

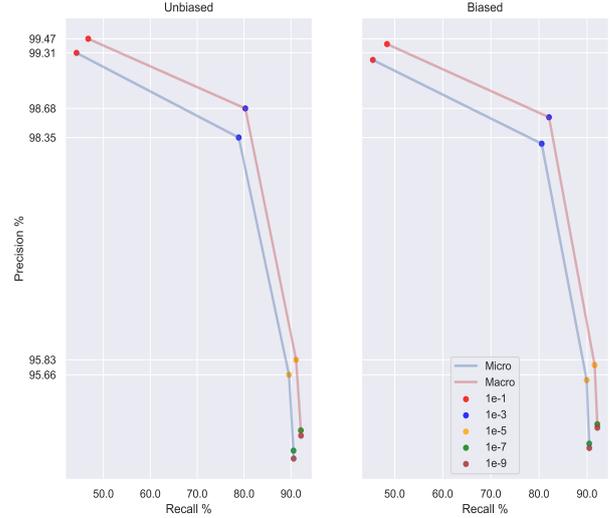

Figure 4: Precision-Recall curve for ParsBERT v3.0 - Real-word Errors.

Next, only the selected candidates were biased using the proposed method, and the best threshold was selected for BERT's score (K). However, it is noteworthy that biasing the model resulted in only marginal improvements (Table 3). We believe this is primarily attributable to the limited volume of generated data used for model biasing.

Figures 3 and 4 illustrate that a lower threshold yields higher recall at the expense of lower precision, while a higher threshold exhibits the opposite trend. Although both 1e-7 and 1e-9 provide higher recall values, the threshold of 1e-5 was chosen as the most suitable. This decision was based on the fact that the discrepancies in precision and recall values between 1e-9, 1e-7, and 1e-5 are minimal, and opting for 1e-5 results in a less stringent model for detecting real-word errors within sentences. This attribute is more desirable in real-world applications, where it is less preferable for the model to introduce errors into correctly spelled words than to overlook real-word errors.

### 5.3. Comparison with the Baseline Method

As mentioned earlier, our research builds on the main idea presented by Jing et al. [36] and proposes a similar method with several notable improvements. Our goal was to enhance the effectiveness and efficiency of the original approach while addressing some of its limitations.

As previously stated, our research is based on the main idea introduced by Jing et al. [36], offering a method that shares some similarities while incorporating several significant enhancements. Our goal was to enhance the effectiveness and efficiency of the original approach while also mitigating some of its inherent limitations.

To compare our biased models to those proposed by Jing et al., we implemented their algorithm and evaluated it using our own dataset. We also used both ParsBERT v2.0 and ParsBERT v3.0 for the BERT model in Jing et al.'s proposed algorithm to ensure that the experiments were conducted under the same and equal conditions. It is worth noting that only unbiased models were used to implement the paper's [36] method, as Jing et al. only used an unbiased BERT model.

In their paper [36], two distinct methods were proposed. The first method involves applying the Levenshtein distance after utilizing BERT to sort the suggestions provided by BERT. Conversely, the second method employs the Levenshtein distance before employing BERT, thereby generating initial candidate suggestions for BERT.

Furthermore, in the first method, a threshold value of "500" was employed to limit the number of expected suggestions from BERT, as this threshold was found to yield superior performance. In the second method, the Levenshtein distance algorithm recommended all words that were similar to the misspelled word and had a distance of 2 or less from the misspelled word.

Now turning our attention to the results obtained from our experiments (as shown in Tables 3 and 4), it becomes evident that our proposed method outperforms the other baseline methods in several key aspects. Table 3 shows a significant improvement in all metrics achieved by our approach compared to that reported in Table 4. This noticeable enhancement can be attributed to our refined algorithm, biasing method and heuristic techniques.

Moreover, the reported evaluation times (as depicted in Tables 3 and 4) highlight the efficiency gains enabled by our method in comparison to Version-2, as proposed in the paper [36]. Through the integration of advanced data preparation techniques and algorithm optimization, we have effectively reduced processing time while significantly enhancing performance. This enhancement holds particular significance in real-time applications where swift decision-making is rec-ommended.

Table 3: Evaluation results for ParsBERT v2.0 and ParsBERT v3.0 with a threshold of 1e-5 (R: Real-word Errors, N: Non-real-word Errors). All values are expressed in percentages, and the time is measured in minutes. This experiment was done on a Nvidia GeForce RTX 3090 with 24 GB RAM.

| Model | ParsBERT v2.0 | | ParsBERT v3.0 | |
| --- | --- | --- | --- | --- |
| Type | Unbiased | Biased | Unbiased | Biased |
| Precision$^{\text{N, Micro}}$ | 79.34 | **79.59** | 76.07 | 75.85 |
| Precision$^{\text{N, Macro}}$ | 79.34 | **79.59** | 76.07 | 75.85 |
| Accuracy$^{\text{R, Micro}}$ | 92.19 | 92.5 | 92.76 | **92.89** |
| Accuracy$^{\text{R, Macro}}$ | 93.56 | **93.81** | 93.57 | 93.78 |
| Recall$^{\text{R, Micro}}$ | 91.34 | **92.16** | 89.57 | 89.92 |
| Recall$^{\text{R, Macro}}$ | 92.76 | **93.54** | 91.09 | 91.58 |
| Precision$^{\text{R, Micro}}$ | 92.91 | 92.79 | **95.66** | 95.60 |
| Precision$^{\text{R, Macro}}$ | 94.25 | 94.05 | **95.83** | 95.77 |
| F1-Score$^{\text{R, Micro}}$ | 92.12 | 92.47 | 92.51 | **92.67** |
| F1-Score$^{\text{R, Macro}}$ | 92.99 | 93.24 | **93.74** | 92.86 |
| Evaluation Time | 123.75 | 123.31 | 100.65 | 101.55 |

Table 4: Evaluation results for Jing et al. [36] paper (R: Real-word Errors, N: Non-real-word Errors). All values are expressed in percentages, and the time is measured in minutes. This experiment was done on a Nvidia GeForce RTX 3090 with 24 GB RAM.

| Model | ParsBERT v2.0 | | ParsBERT v3.0 | |
| --- | --- | --- | --- | --- |
| Version | Version-1 | Version-2 | Version-1 | Version-2 |
| Precision$^{\text{N, Micro}}$ | **68.11** | 67.46 | 65.19 | 60.47 |
| Precision$^{\text{N, Macro}}$ | **68.12** | 67.46 | 65.19 | 60.47 |
| Accuracy$^{\text{R, Micro}}$ | **74.05** | 72.39 | 73.87 | 64.01 |
| Accuracy$^{\text{R, Macro}}$ | 74.66 | 72.69 | **77.38** | 64.76 |
| Recall$^{\text{R, Micro}}$ | 58.08 | **73.42** | 59.69 | 65.18 |
| Recall$^{\text{R, Macro}}$ | 57.98 | **74.08** | 65.42 | 66.13 |
| Precision$^{\text{R, Micro}}$ | **85.32** | 71.94 | 83.33 | 63.68 |
| Precision$^{\text{R, Macro}}$ | **86.48** | 72.06 | 84.73 | 64.35 |
| F1-Score$^{\text{R, Micro}}$ | 69.11 | **72.67** | 69.56 | 64.42 |
| F1-Score$^{\text{R, Macro}}$ | 65.12 | **72.94** | 67.51 | 64.85 |
| Evaluation Time | 45.5 | 249.5 | 45 | 245 |

However, it is worth noting that Version-1, is faster than our approach, primarily due to the simplicity of the algorithm employed in that version. It is evident that prioritizing speed can lead to trade-offs in other metrics, such as precision and recall. Nonetheless, when speed is of paramount importance, opting for a simpler version is advisable.

In conclusion, through extensive experimentation and comparative analyses, it is demonstrated that our approach surpasses the other methods in [36] in terms of accuracy, precision, recall, and computational efficiency (compared to Version 2). These findings highlight the advancements made in this field and emphasize the potential impact of our proposed method on future research and practical applications.

### 5.4. Comparison with External Systems

This section involves a comparative assessment of the models' performance in relation to several established systems. To facilitate this evaluation, a set of 100 sentences, each containing an error, was carefully selected. The systems' and our models' capabilities in correcting various types of errors were evaluated in this context. These 100 sentences were individually assessed, and the efficiency of the different methods was measured by counting the number of corrected errors. The reason for limiting the evaluation to 100 sentences is the necessity for manual inspection of the sentences in the two systems introduced, which is a time-consuming operation.

The comparative analysis is specifically conducted against two Google systems: the Google Translate system[1] and the Gmail system[2]. It is essential to underscore that the principal objective of these two systems is not error correction; they offer this functionality as an ancillary feature. More precisely, in the Google Translate system, the evaluation is based on the text translation output and the suggestions provided under the "Did you mean..." blue line. In the case of the Gmail system, its spell-checking feature is employed to assess its performance.

Figure 5 illustrates that our models exhibit superior performance compared to the other two systems. Notably, our models demonstrate a significantly greater capacity to correct real-world errors. However, it should be acknowledged that the Google Translate model outperforms our models when it comes to correcting non-real-word errors.

In summary, the results of the comparative analysis between our models, Gmail, and Google Translate reveal that our models excel in detecting a substantially higher number of errors.

## 6. Discussion

### 6.1. Error Analysis

In order to evaluate the strengths and weaknesses of the models, each metric was computed for distinct error types individually. As depicted in Figure 6, it is evident that the models exhibit similar performance, with variations typically within the range of 1 ∼ 2%, across different categories of non-real-word errors.

---

[1] https://translate.google.com
[2] https://mail.google.com

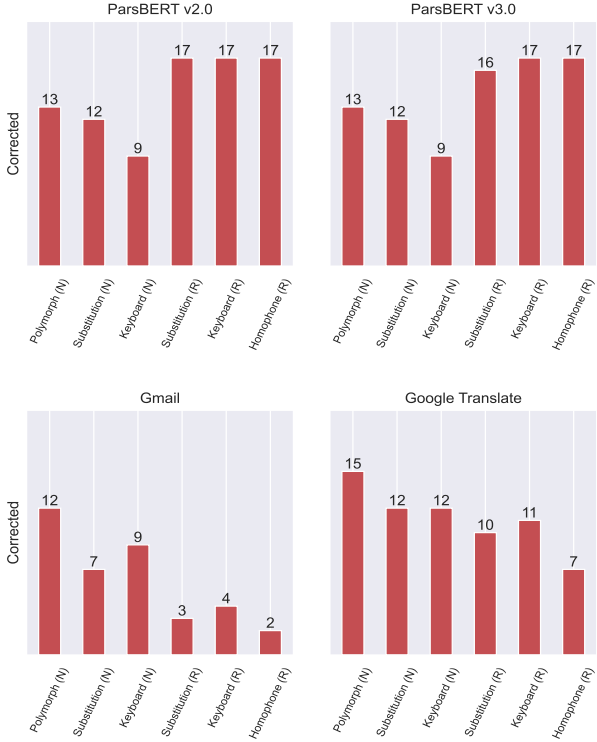

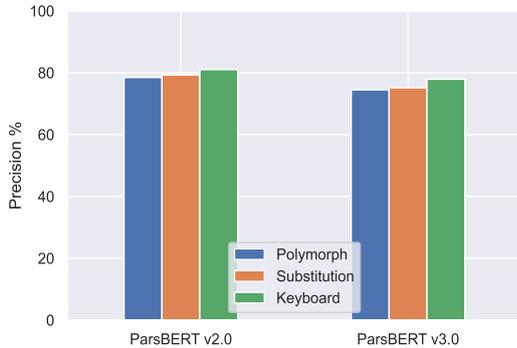

Figure 5: The performance of biased models with threshold 1e-5, Gmail, and Google Translate on 100 sentences. (R: Real-word Errors, N: Non-real-word Errors)

Figure 6: Precision of biased models on each non-real-word error types.

Table 5: The F1-Score of biased models, with a threshold of 1e-5, for each category of real-word errors. All values are expressed in percentages.

| Error Type | ParsBERT v2.0 | ParsBERT v3.0 |
|---|---|---|
| Homophone | 97.45 | 96.22 |
| Substitution | 92.57 | 91.49 |
| Keyboard | 89.69 | 90.88 |

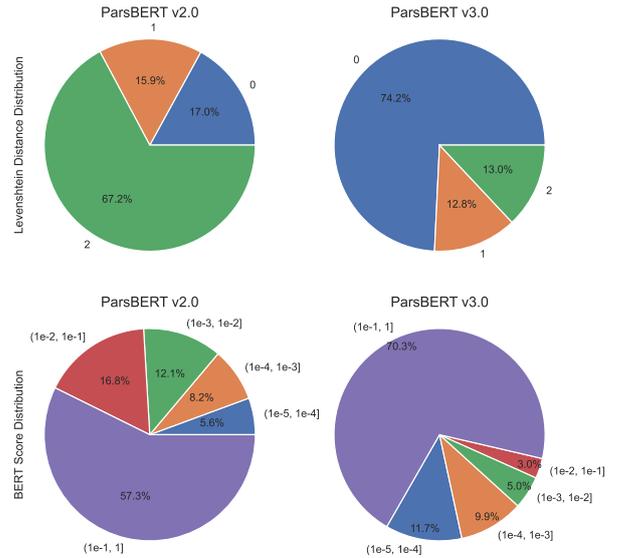

Figure 7: The distribution of Levenshtein distance and BERT scores for missed keyboard real-word errors by biased models with a threshold of 1e-5.

Conversely, the models exhibited notably stronger performance when addressing real-word errors, as highlighted in Table 3. However, it's important to note that certain error types posed more significant challenges to the models than others. For instance, Table 5 shows the models' effectiveness in rectifying homophone real-word errors. Nevertheless, the outcomes for substitution errors, and particularly keyboard real-word errors, suggest that these error categories were more challenging to correct due to their inherent complexity.

For a more detailed analysis, Figure 7 provides insight into the distribution of Levenshtein distance and BERT scores for a scenario in which the model missed keyboard real-word errors. In other words, input is a keyboard real-word error and output is a wrong suggestion. Notably, the majority of incorrect suggestions by ParsBERT v2.0 have a Levenshtein distance of 2 from the input words, indicating that this model struggled to select the most appropriate suggestion from the data structure outlined in Section 4.6. However, it will be discussed (See Section 6.2) that ParsBERT v2.0 cannot handle Zero-width Non-joiner Space (ZWNJ); therefore, majority of errors were caused because of ZWNJ. In the case of ParsBERT v3.0, most errors occurred when the model suggested the exact same output as the input word. This demonstrates that the model was unable to detect most of the errors initially.

Regarding the BERT scores assigned to incorrectly suggested words, it is noteworthy that more than half of the errors in both models had scores falling within the range of 0.1 to 1.0, which is generally considered a good score for a suggestion. Therefore, the primary issue appears to be related to the BERT model's challenge in accurately scoring suggestions, which, in turn, leads to the models' incorrect recommendations.

### 6.2. Zero-width Non-joiner Space in Persian

The Zero-width Non-joiner (ZWNJ) space in Persian is a type of space that is inserted between certain letters and words.

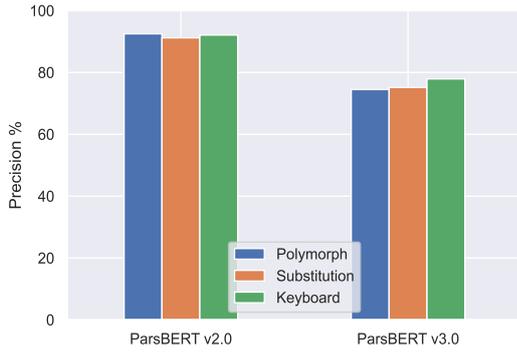

Figure 8: Precision of biased models on each non-real-word errors with replaced ZWNJ space.

It appears as a small, very thin vertical line. ZWNJ space serves to enhance text readability by increasing the spacing between specific letters and words. For instance, in the phrase "جمله‌ها" (meaning "Sentences"), the words "جمله" ("Sentence") and "ها" (a postfix to make plural nouns) are separated by a ZWNJ space. Consequently, writing it as "جمله ها" or "جملهها" is deemed incorrect in formal writing.

Generally, the use of a ZWNJ space in Persian enhances text readability and comprehensibility. It is widely employed in typing and writing texts. Nevertheless, it may pose challenges for model performance if the BERT model lacks training on words containing ZWNJ spaces. ParsBERT v3.0 stands out as a model capable of handling words with ZWNJ spaces, thereby addressing the ZWNJ space issue. Consequently, despite demonstrating similar results on the evaluation dataset, ParsBERT v3.0 is the preferred model for real-world applications.

To draw a more definitive conclusion, a scenario was simulated in which ZWNJ spaces were eliminated from both the labels and the models' suggestions (by replacing all ZWNJ spaces with empty characters). The performance of both models was then compared under these conditions.

The simulation results revealed that ParsBERT v3.0's performance remained unchanged, as anticipated. In contrast, ParsBERT v2.0 exhibited a noteworthy improvement in performance (refer to Figures 6 and 8). A similar trend was noted for real-world errors, with each metric for ParsBERT v2.0 showing an increase of at least 3 ∼ 4%.

In addition, Figure 9 has been recreated (based on Figure 7) to illustrate the impact of ZWNJ spaces in keyboard real-word errors. The distribution of both Levenshtein distance and BERT scores for ParsBERT v2.0 has been altered. When comparing Figures 7 and 9, it becomes evident that, with the removal of ZWNJ spaces, a significant number of errors with a Levenshtein distance of 1 have been corrected. However, errors with a distance of 0 remained unchanged, although their overall count increased (this percentage increase is attributed to the reduction in the total number of incorrect suggestions). Furthermore, a majority of errors with a Levenshtein distance of 2 were transformed into errors with a distance of 1.

Overall, all these results underscore that ParsBERT v2.0 is unable to handle ZWNJ spaces and would outperform ParsBERT v3.0 in the absence of ZWNJ spaces.

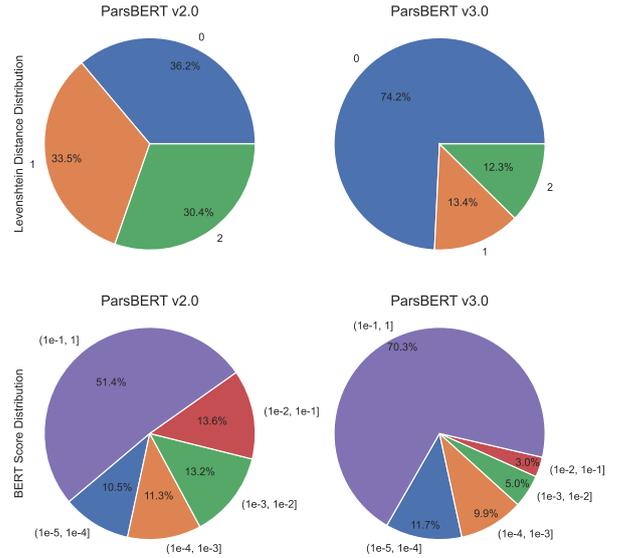

Figure 9: The distribution of Levenshtein distance and BERT scores for missed keyboard real-word errors by biased models with a threshold of 1e-5 with replaced ZWNJ space.

## 7. Conclusions

In this research, we presented a comprehensive algorithm for addressing spelling errors using the BERT masked language model. The evaluation results showcased the effectiveness of our models and algorithm in rectifying spelling errors.

Furthermore, our results benefited from fine-tuning pre-trained BERT models on our dataset, leading to improved performance. Notably, our proposed method demonstrated greater robustness and potency when compared to the baseline approach of using BERT for spelling correction.

In a broader context, while we endeavored to address various aspects, there is room for improvement. Regrettably, model performance can be impacted by short sentences and multiple spelling errors within a single sentence. Therefore, the development of a model capable of correcting a larger number of words within a sentence represents an important avenue for future research.

To conclude, this project addresses a specific subset of writing issues, leaving more extensive concerns such as automatic paragraphing, punctuation checking, and grammar rules unexplored. The development of a comprehensive system encompassing all these aspects, akin to Grammarly[1] for the Persian language, holds potential for practical and effective outcomes.

## 8. Acknowledgments

We would like to express our gratitude for the partial support provided by DeepMine, an AI company in Iran. Addition-ally, we extend our thanks to Taaghche application for generously providing unlabelled text data for our training purposes. Your contributions have been invaluable to the success of this research.

---

[1] https://app.grammarly.com

All source codes and materials are available at this link[1].

---

[1] https://github.com/Amir79Naziri/SpellCorrectionML